Submitted as part of the
*Roadmap on Soft Robotics:*
*multifunctionality, adaptability and growth without borders*
to be published in IOP's
*Multifunctional Matetials*

# Section 06 – Sustainably Grown: The Underdog Robots of the Future
Stoyan K. Smoukov, Active and Intelligent Materials Lab, Queen Mary University of London, UK

**Status**
Automata have captured people's imagination for millennia. From Ctesibius' water clock and moving owl more than two centuries B.C., recent integration with electronics and independent high density power sources has produced today's mobile robots. These including rescue robots resembling dogs, snakes, and other animals. In humanoids, the last two decades' exponential developments have eclipsed the 2002-2004 humanoid RoboCup where just standing in place and kicking a ball was considered the top achievement. The recent demos by Boston Dynamics[1] of robots rebalancing after being pushed, as well as synchronized dancing skills, show the goal of a RoboCup team of robots beating the human world champion team by 2050 by FIFA rules is very much on track.

It is hard to imagine current approaches are lacking somewhere, yet they will not be applicable to the majority of robots in the near future. We are on the verge of two new transitions that will transform robotics. One is already under way - the miniaturization of robots, to the point where invisible, microscopic robots could be around us and inside us, performing monitoring or even life-saving functions. We have seen systematic bio-inspired efforts to create microbe-like, microscopic robots. The trend has parallels with miniaturization in the electronics industry, where exponentially smaller and more energy efficient units have been produced each generation. To put this statement in context, examples already include magnetic microswimmer robots, employing bacterial modes of locomotion, which are biocompatible, potentially ready for integration within our bodies. They require lithography to create clever microscopic screw-type structures,[2] enough to produce the cork-screw swimming movement. Such micro-robots have encapsulated, picked, and delivered cells, protecting them from shear forces in fluids,[3] while others have captured non-motile sperm, propelled them, and ultimately fertilized an egg.[4] We explore how such developments in micro-robots will change our world in the relatively near future.

The second trend is bottom-up robotics, growing robots from a solution medium, as if they were bacteria. This field is emerging at the intersection of a number of disciplines, discussed below. An overarching common theme is the creation of artificial life from a non-biological starting point.

**Current and Future Challenges**
Sustainability and energy efficiency are front-and-centre issues as robotics becomes more useful, more affordable, and therefore more ubiquitous. This is especially true of microscopic robots that would be too small and cheap to collect and recycle traditionally. All current microelectronics and most micro-robots rely heavily on lithography for their manufacturing. This top-down method, though extremely precise, is rather wasteful. For highly integrated chips, a 20 g chip requires 1.7 kg of materials inputs.[5] Bottom-up sustainable approaches can mitigate/eliminate such waste.
E.g., inspired by plant roots, a robot created an embedded "root" system building a tube around a 3D printing head that also doubled as a burrowing tip.[6] This creative additive strategy may greatly decrease the environmental impact and increase the versatility of architecture of embedded tube networks down to millimeters. New physical principles and interactions between building components are needed, however, for effecting growth of robots on the micro- and nanoscale to parallel that in biological cells and cellular components. The aim would be to "grow" robots at the same cost as plants and bacteria, with a similar end-of-life decomposition cycle.

As micro-robots get smarter, a natural impulse is to integrate them with the electronics which we associate with computation. Smarter micro-robots could greatly enhance the field of electroceuticals,[7] e.g. with their ability to relocate to different nerves and sense/deliver the right stimuli. It is also worth keeping in mind that in nature the most energy efficient computation devices, our brains, run on ionics, not electronics. Since 2007, huge gains in the field of ionic diodes[8] promise the creation of ionic logic and computation.

Creating artificial life is another blue sky challenge that may be achieved within the lifetimes of some present readers. On the small scale numerous challenges include how to harvest energy and integrate power systems, defining shape, enabling locomotion, environmental sensing, computation, enabling division, and eventually reproduction, heredity, and evolution. In this respect the goals of engineering robotics on the microscale overlap with those of synthetic biology in its purest form. In the next section we show significant progress in almost all the above areas starting from abiotic compounds.

**Advances in Science and Technology to Meet Challenges**
The smallest articulated swimmer (just <1μm) is limited not only by the expense of lithography, but also by the inadequacy of simply miniaturizing large structures. New principles needed for movement at smaller length scales also afford simpler designs. Understanding self-assembly and phase behaviours of small molecules would let us grow them as plants and animals grow in nature, from nutrient media using internal programmed mechanisms. Materials science and robotics are currently in a stage of development of learning from life at those scales.[9]

The discovery of artificial morphogenesis[10] from cooling, has allowed spherical oil droplets to transform in both sizes[11] and into shaped polymer particles.[12] Phase transitions yield regular 3D (icosahedra and octahedra) and quasi-2D geometric shapes (hexagons, pentagons, rhomboids, triangles, trapezoids and rods). Only two chemical components are needed besides water, to grow such shapes, and remarkably only two components also for producing flagellated active swimmers, which can be recharged from fluctuations in the environment.[13] Further functional integration in synthetic biology would let us mimic cells with minimal number of components.[14]

A combinatorial approach to multifunctional polymers has recently removed the unpredictability of including extra functions in single molecules by interpenetrating multiple phases with their separate functions. Both simultaneous synthesis and sequential interpenetration are possible. [15] Combining ionic actuation and shape memory has achieved programmable movement[16], supercapacitor energy storage,[17] and self-sensing muscles[18]. All these techniques are compatible with efforts under way to combine multi-functionality with bottom-up shaped polymer particles, to grow swimming, self-sensing, programmable and autonomous polymer robots. Self-healing is being developed for materials to drive energy harvesting and storage in the next generation of micro-robots.[19]

Peptide amphiphiles are now assembling gels and tissues with biological signalling and nanofiber materials with state-of-the-art tunable functionality. Recent structures with both covalent and non-covalent interactions expand greatly the ability for tunability, responsiveness and repair, as well as dynamic functional assemblies. [20] Sequence-defined polymers are aiming to play the role of proteins in new assembly and nano-robotic systems. Progress is being made in error-free sequence synthesis, reading sequences, and also developing the potential for repair and the creation of catalytically active analogues of enzymes.[21]

**Concluding Remarks**
The drive of micro-robotics towards artificial life is both fundamental and practical. Constructing a living system would be the most concrete possible answer to the Origin of Life, while also exploring minimum requirements and expanding the conditions under which it could thrive. Practically, a population of micro-robots could be persistently maintained under the right circumstances, though individuals die and decompose, consistent with the goals of sustainability.

Growing invisible microscopic robots bottom-up is the convergent goal of several different fields – from self-assembly of responsive materials, to bottom-up synthetic biology,[22] to physics and chemistry of active matter. They all aim at reproducing aspects of life, both for understanding life's mystery, and re-creating life "from scratch", as "doing" is the ultimate form of "understanding". Reproduction in life is also the ultimate form of self-repair – with the promise of improvement and adaptation from generation to generation.

**Acknowledgements**
Acknowledging funding from the UK EPSRC fellowship (EP/R028915/1).